%% file: ICCV2025-Author-Kit-Feb/main.tex
\documentclass[10pt,twocolumn,letterpaper]{article}

\usepackage[pagenumbers]{ICCV2025-Author-Kit-Feb/iccv}      % To produce the REVIEW version

\definecolor{iccvblue}{rgb}{0.21,0.49,0.74}
\usepackage[pagebackref,breaklinks,colorlinks,allcolors=iccvblue]{hyperref}
\usepackage{dsfont}
\usepackage{rotating}
\usepackage{bm}
\usepackage{placeins}
\usepackage[linesnumbered,ruled,vlined]{algorithm2e}
\usepackage[dvipsnames]{xcolor}
\usepackage{amsthm}
\usepackage{amsmath}
\theoremstyle{plain}
\newtheorem{theorem}{Theorem}[section]
\newtheorem{proposition}[theorem]{Proposition}
\newtheorem{lemma}[theorem]{Lemma}

\theoremstyle{definition}

\theoremstyle{remark}

\def\paperID{*****} % *** Enter the Paper ID here
\def\confName{ICCV}
\def\confYear{2025}

\title{TAPS : Frustratingly Simple Test Time Active Learning for VLMs}

\author{Dhruv Sarkar$^{1}$\thanks{All three authors contributed equally.} \ \ \ Aprameyo Chakrabartty$^{1}$\textsuperscript{*} \ \ \ Bibhudatta Bhanja$^{1}$\textsuperscript{*}\\
$^{1}$IIT Kharagpur\\
\small{\texttt{\{dhruv.sarkar@kgpian.,aprameyo@kgpian.,bbhanja@kgpian.\}iitkgp.ac.in}}}

\begin{document}
\maketitle
\input{ICCV2025-Author-Kit-Feb/sec/0_abstract}    
\input{ICCV2025-Author-Kit-Feb/sec/1_intro}
\input{ICCV2025-Author-Kit-Feb/sec/2_method}

\input{ICCV2025-Author-Kit-Feb/sec/3_theoreticaljustification}
\input{ICCV2025-Author-Kit-Feb/sec/4_experiment}
\input{ICCV2025-Author-Kit-Feb/sec/7_conclusion}
\iffalse
{
    \small
    \bibliographystyle{ieeenat_fullname}
    \bibliography{ICCV2025-Author-Kit-Feb/main}
}
\fi

\input{ICCV2025-Author-Kit-Feb/main.bbl}
\input{ICCV2025-Author-Kit-Feb/sec/X_suppl}

\end{document}

%% file: ICCV2025-Author-Kit-Feb/sec/0_abstract.tex
\begin{abstract}
Test-Time Optimization enables models to adapt to new data during inference by updating parameters on-the-fly. Recent advances in Vision-Language Models (VLMs) have explored learning prompts at test time to improve performance in downstream tasks. In this work, we extend this idea by addressing a more general and practical challenge: Can we effectively utilize an oracle in a continuous data stream where only one sample is available at a time, requiring an immediate query decision while respecting latency and memory constraints? To tackle this, we propose a novel Test-Time Active Learning (TTAL) framework that adaptively queries uncertain samples and updates prompts dynamically. Unlike prior methods that assume batched data or multiple gradient updates, our approach operates in a real-time streaming scenario with a single test sample per step. We introduce a dynamically adjusted entropy threshold for active querying, a class-balanced replacement strategy for memory efficiency, and a class-aware distribution alignment technique to enhance adaptation. The design choices are justified using careful theoretical analysis. Extensive experiments across 10 cross-dataset transfer benchmarks and 4 domain generalization datasets demonstrate consistent improvements over state-of-the-art methods while maintaining reasonable latency and memory overhead. Our framework provides a practical and effective solution for real-world deployment in safety-critical applications such as autonomous systems and medical diagnostics.
\end{abstract}

%% file: ICCV2025-Author-Kit-Feb/sec/1_intro.tex
\section{Introduction}
\label{sec:intro}

%``\textit{Simplicity is the ultimate sophistication}"

%\hspace{50mm} -Clare Boothe Luce
%\newline\newline
The emergence of large Vision-Language Models (VLMs) \cite{radford2021learning,li2022blip,fini2023improved} has revolutionized visual understanding by enabling zero-shot generalization through vision-language alignment. While prompts like ``a photo of a {class}" guide VLM's text-visual comparisons, crafting optimal prompts remains heuristic-driven, spurring parameter-efficient prompt learning \cite{zhou2022learning,wang2023position} that tunes soft prompts without modifying pretrained models. Additional background on Prompt Learning and VLMs is provided in section \ref{sec:preliminaries} of the appendix for interested readers.

While supervised prompt learning \cite{zhou2022conditional,khattak2023maple} achieves strong discriminative performance, its practical limitations become apparent when facing real-world deployment challenges: annotation scarcity in specialized domains (e.g., rare medical pathologies), dynamic environments requiring continuous adaptation (e.g., autonomous driving), and latency-critical applications prohibiting offline retraining. Contemporary Test-Time Adaptation (TTA) approaches \cite{wang2020tent,yuan2023robust} circumvent these barriers through self-supervision rather than pure unsupervised learning—employing surrogate training signals like entropy minimization over test-time augmentations \cite{shu2022test} or feature distribution alignment via moment matching \cite{liang2024comprehensive}. Nevertheless, such unsupervised adaptation mechanisms frequently exhibit performance disparities relative to fully supervised baselines. To mitigate this limitation, active learning methodologies \cite{ren2021survey,zhan2022comparative} offer a hybrid solution, strategically curating instances for annotation to enhance predictive performance with minimal labeling expenditure.

TAPS (\textbf{T}est-Time \textbf{A}ctive \textbf{P}rompt Learning for \textbf{S}ingle-Sample Streams) is a novel generalization framework that dynamically identifies uncertainty in model predictions, triggering oracle annotation requests for critical samples. Upon label acquisition, the system jointly optimizes prompts using both annotated instances and the broader unlabeled test distribution. The principal innovation lies in addressing the non-trivial challenge of active sampling under streaming data constraints—specifically, operating on individual data points rather than batched inputs. This work represents the inaugural effort to address active sampling within a continuous data stream paradigm, processing individual instances sequentially while maintaining real-time adaptation capabilities. The method is frustratingly simple but keeps both latency and memory constraints in mind. Further, the simplicity of the design choices makes them amenable to conceptual and mathematical justification. We believe that as a pioneering work in this area, we should start with something that is simple yet has strong conceptual foundations.

\paragraph{Motivation, Challenges and Novelty.} Models are often deployed in many safety-critical scenarios. In such scenarios, it is often better to consult an oracle and learn from its advice rather than taking the risk of wrong prediction. However, as this is essentially a test-time setting, we have to keep certain things in mind-
\begin{itemize}
    \item The decision to consult an oracle cannot be indefinitely postponed. This is because in many real life situations, the expert feedback is needed instantaneously, like in the case of a self-driving car or other autopilot systems.
    \item There cannot be a presupposition of the data arriving in a batched manner - that is, we have to work with the assumption of only a single sample arriving. The assumption of multiple test points being present at a time is restrictive and could prevent widespread use of the model\cite{zhang2022memo}. It is also not wise to wait for the arrival of multiple samples to form a batch.
    \item The latency of the system cannot be too high as this is a test-time setting.
    \item The memory requirements of the system should be limited.
\end{itemize}
Incorporating these characteristics in a methodology would make it highly applicable but significantly more challenging than regular active learning applications. This is because -
\begin{itemize}
    \item Making the query decision at the time of arrival of the sample in real-time would mean that we are doing so without knowing what the future data stream would look like. Under query budget constraints, this becomes an important factor where, ideally one should prioritise the most uncertain samples for labelling. Thus, any such decision-making mechanism would be fundamentally heuristic.
    \item In a batch, one can choose the the most uncertain samples of the batch. This is a reasonable method and has been tried in various works\cite{gui2024active} before. However, this method won't work in our case as we only have a single sample at a time.
    \item The active learning mechanism should be simple yet effective - more complicated mechanisms that could yield slightly better results while significantly compromising latency are not acceptable - while the former is important, the latter is equally important in a test-time setting.
    \item There is only limited time to learn from labelled samples as they cannot be held on to for a long time due to memory constraints.
\end{itemize}
We come up with a method that is simple, but non-trivially takes into account all the above considerations. Thus, while we have demonstrated our work on VLMs due to their ubiquity, it must be appreciated in its full generality. It is our strong belief that this general framework would spur on further developments in this nascent field.
\paragraph{Comparison with prior works.}
A contemporary work ATTA \cite{gui2024active} shows the impact of active learning in the standard TTA setting. Their analysis established more generalization ability in the model than standard TTA setting. However, their methodology and framework cannot be directly applied to large VLMs like CLIP \cite{radford2021learning} because they do not focus on the adaptation of a specific parameter group in a large model, while ours does so by incorporating the parameter-efficient Prompt Tuning. Further, they assumed a batch setting whereas we have a single test sample at a time and they also did multiple gradient updates per minibatch while we have only one gradient update per sample. As our approach updates and improves the model after encountering a single test example without waiting for additional examples to arrive, it is more flexible in continuous data streaming scenarios \cite{zhang2022memo}. Further, relying on the batch setting, they use a complicated clustering mechanism which while effective, has great latency cost and greatly limits the applicability of this framework to test time scenarios. At the same time, it is much more challenging to select samples to query and also to ensure that budget is not exceeded.
There is another recent work \cite{bang2024active} which has used Prompt Learning and Active Learning together, but that was not in a test-time setting let alone one with only one sample at a time (which makes our problem significantly more challenging), and they also did not try on adaptation tasks, so a direct comparison with it is simply not feasible. Readers are requested to check section \ref{app:related work} of the appendix for more details on related works.
\newline\newline
Our contributions are the following-
\newline
\begin{enumerate}
    \item To the best of our knowledge, we are the first to apply active learning in a test time setting with a test stream of only one sample at a time while not postponing the query decision. To instantiate our approach we use VLMs, which have the advantage of being widely applicable while having well established methods of parameter-efficient and low-latency tuning methods such as prompt tuning.
    \newline
    \item We use the innovative idea of using a dynamically adjusted entropy threshold to decide which test time samples have to be queried. We also incorporate the idea of class balancing in the annotation buffer and the replacement of non-informative samples which ablative studies reveal to be crucial to our performance.
    \newline
    \item For adaptation tasks, we also use the novel idea of class aware distribution alignment which makes effective use of the actively labelled samples to achieve more fine-grained distribution alignment.
    \newline
    \item The proposed algorithm performs better than many other methods under a low annotation budget and a limited buffer capacity. The evaluation protocol under which it was evaluated is demonstrably fair.
\end{enumerate}
\begin{figure*}[h]
  \centering
  \includegraphics[width=\textwidth]{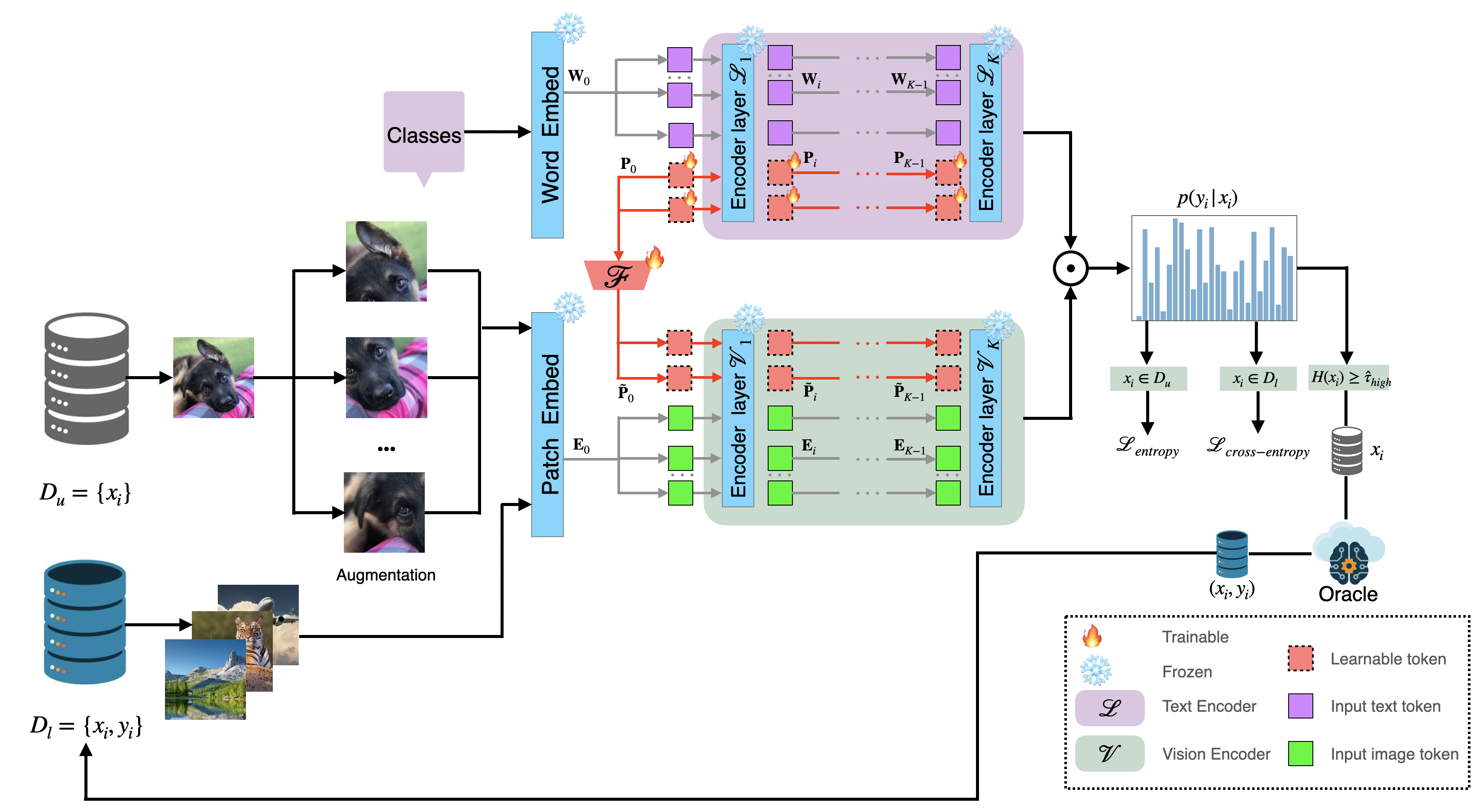}
  \caption{Overview of our method. We query uncertain samples at test time and put them in a buffer of limited size. We decide to query a sample if its entropy is above a certain threshold ($\tau_{\text{high}}$) which is dynamically adjusted. The old samples of a disproportionately represented class in the buffer are removed when they are not informative anymore.}
  \label{fig:Figure 1}
\end{figure*}

%% file: ICCV2025-Author-Kit-Feb/sec/2_method.tex
\section{Method}
\label{sec:formatting}

\begin{table*}[h!]
\caption{Motivating findings of not updating on low confidence / top 1\% highest entropy (dynamically estimated) samples}
\centering
 \begin{tabular}{||c | c |c ||} 
 \hline
 Dataset & CLIP+TPT & CLIP+TPT, not updating on top 1\% \\\hline
  DTD & 47.75 & 47.80 \\
  Stanford Cars & 66.87 & 66.93 \\
  Oxford Pets & 87.79 & 87.85  \\
    Food101& 84.67 & 84.70  \\
  \hline
 \end{tabular}
\end{table*}
One shortcoming with the approach in TPT is that in trying to increase the certainty and consistency of a test sample across views, it potentially learns spurious representations. This is due to the presence of samples for which it is generally uncertain, and imposing consistency and certainty on such samples will potentially make the model predict wrong labels with certainty. We confirm this hypothesis by doing an experiment where we do not update on those samples with high entropy and instead just evaluate on them and let them pass. The results are given in Table 1. The improvement may seem marginal, but it must be noted that this was just a toy experiment that simply involved an act of omission, not one of commission. Thus, the results must be seen in that light. Simply ommiting the highly uncertain samples gives us an improvement, even if marginal, this begs the question - can we make further use of the highly uncertain samples in a better way by employing active learning so as to not waste the information they contain?
To mitigate the learning of spurious representation via updation on highly uncertain samples and also to not waste the information that could be provided by them by removing them from the training scheme, we propose to actively query some samples for which the model is generally uncertain. We formalise this notion of general uncertainty by selecting those samples for which the entropy of the average logit is more than a certain threshold which is \textit{dynamically} adjusted. Our approach is visualised in Figure \ref{fig:Figure 1}.
We tried a fair evaluation protocol, which we have described in section \ref{eval_protocol}. The most crucial detail in an evaluation protocol is the point in time at which we solicit the true label from the oracle. In particular, when we solicit the label before evaluating the sample, we cannot have a fair evaluation. Thus, our evaluation makes sure we are not using the ground truth label from the oracle and still keeping the sample for evaluation. In samples that are not queried, we apply the standard unsupervised loss as marginal entropy of the average logit.
\subsection{Formalisation}
We have a buffer $\mathcal{D}_l$, which is initially empty and is used to store queried samples. The buffer size is assumed to be limited to ensure that the framework is realistic. Let $\mathcal{D}_u$ denote the entire dataset. At time $t$, a test sample $x_t \in \mathcal{D}_u$ arrives to be evaluated. The test sample is augmented $N$ times to produce a batch of augmentations $\Tilde{x}_t$. Like TPT, we find the logits for each of them and calculate the marginal entropy. The noisy augmentations are thus discarded. The average of the remaining logits is computed and the entropy corresponding to that is found. We denote that entropy as $H(x_t)$. If $H(x_t) > \tau_h$, then we choose to query it. 
\newline \newline
\iffalse
\begin{algorithm}[htbp]
    \caption{Dynamic Threshold Selection}
    \label{alg:Algorithm 1}
    
    \KwIn{$t$, $N_{queried}$, $\hat{\mu}_t$, $\hat{\sigma}_t$}
    
    \eIf{$t < T_{min}$}{
        \textcolor{Orchid}{\#Initially select a static threshold till $\tilde{t}$ samples}\;\vspace{-0.2cm}
        \begin{flushleft}
            \KwOut{$\tau_0$}\;
        \end{flushleft}
        \vspace{-0.2cm}
    }{
        \textcolor{Orchid}{\#$\alpha$ is the query budget percentage}\\
        \eIf{$\frac{N_{queried}}{t} \geq \alpha$}{
            \textcolor{Orchid}{\#Select higher value of $z$ ($z_{high}$) if currently over-querying}\;
         
            $
            \tau_t = \hat{\mu}_t + z_{high} \cdot \hat{\sigma}_t
            $\;
        }{
            Select standard value of $z$ with respect to $\alpha$ otherwise
            $
            \tau_t = \hat{\mu}_t + z_{selection} \cdot \hat{\sigma}_t
            $\;
        }
        \KwOut{$\tau_t$}
    }

\end{algorithm}
\fi

\SetKwComment{Comment}{\hspace{0.2em} \# }{}
\SetCommentSty{textnormal}

\begin{algorithm}[htbp]
    \caption{Dynamic Threshold Selection}
    \label{alg:Algorithm 1}
    
    \KwIn{$t$, $N_{\text{queried}}$, $\hat{\mu}_t$, $\hat{\sigma}_t$}
    
    \eIf{$t < T_{\min}$}{
        \Comment{Initially select a static threshold till $\tilde{t}$ samples}
        \KwOut{$\tau_0$}
    }{
        \Comment{$\alpha$ is the query budget percentage}
        \eIf{$\frac{N_{\text{queried}}}{t} \geq \alpha$}{
            \Comment{Select higher value of $z$ ($z_{\text{high}}$) if currently over-querying}
            $\tau_t \leftarrow \hat{\mu}_t + z_{\text{high}} \cdot \hat{\sigma}_t$
        }{
            \Comment{Select standard value of $z$ with respect to $\alpha$}
            $\tau_t \leftarrow \hat{\mu}_t + z_{\text{selection}} \cdot \hat{\sigma}_t$
        }
        \KwOut{$\tau_t$}
    }
\end{algorithm}

\textbf{Dynamic Threshold Selection.}  As explained in the challenges, we do not have access to the data apriori and in such a continuous data-streaming scenario, we have use some form of online estimation to decide upon the entropy threshold which would help us decide which samples to query. The threshold $\tau$ beyond which samples are chosen for active labeling is dynamically adjusted. Let us denote the threshold at time step t as $\tau_t$. 
\begin{equation}
    \tau_t = \hat{\mu}_t + z\hat{\sigma_t}
\end{equation}
Where $\hat{\mu}_t$ and $\hat{\sigma_t}$ denote the estimated mean and standard deviation of $H(x_t)$ upto time step $t$.
For the first $\tilde{t}$ time steps, we keep the threshold static which essentially serves as a regularization mechanism in the sense that it prevents the threshold from wildly fluctuating too early into the data stream. That is, $\tau_t = \tau_0 $ $\forall t \in [\,\tilde{t}\,]$. 

\textbf{Preventing premature exhaustion of budget.} Since we also want to be within our budget and not exhaust it too early in the test stream, we adjust the value of $z$ depending on the proportion of samples queried in the test stream up to that point. That is, if our average query rate is above the ideal query rate (which depends on budget), we increase the value of $z$ which in our case was the z-score corresponding to top 5-percentile to $z_{high}$ which is a specific higher value of z-score, which in our case was the z-score corresponding to top 2.5-percentile, which makes our mechanism more selective. We do the detailed analysis to show that is effective in section \ref{pf:query_adaptive}. Our approach is summarized in Algorithm \ref{alg:Algorithm 1}. The selection policy is effective - as demonstrated in section \ref{ablation:selection}.

%\label{class balancing}
\textbf{Class Balanced Replacement Policy.} \label{class balancing} To model the memory constraints that frequently occur in real-life scenarios, we make use of a limited size buffer to store the actively labelled samples. As our buffer size is limited and the total number of queried samples typically far exceeds that, we came up with a replacement policy which would help us choose which sample to remove from the buffer in favour of an incoming sample. In evicting samples from the buffer, we adopt a notion of diversity achieved via class balancing. That is, we ensure that all classes are well represented in $\mathcal{D}_l$. We remove the sample with the lowest cross-entropy loss in the class with the most number of samples. Suppose there is more than one class with maximum number of samples. In that case, we choose the class with minimum average cross-entropy for tie-breaking.  The rationale is based on sound intuition. The class with the most number of samples is over-represented, and the sample with the least cross-entropy loss is not that informative anymore.  Our approach is summarised in Algorithm \ref{alg:Algorithm 2}.
\paragraph{Class Aware Distribution Alignment.} Previous works\cite{sun2016deepcoralcorrelationalignment,abdul2024align,zellinger2019centralmomentdiscrepancycmd} have utilised a form of distribution alignment in adaptation tasks to bridge the disrepancy between the source and target distributions. We also incorporate the same in our adaptation tasks, while adding a novel modification which further makes use of the unique information provided by actively labelled samples.That is, instead of just aligning the sample statistics with the general (coarse-grained) source statistics we align them to class-statistics (fine-grained) of the source dataset as we already know the class of the actively labelled samples. The idea is that this should lead to more precise positioning of features in the feature space. This shows observable improvement in our results as shown in \ref{sec:dist-align}.
In our case we take the source dataset to be the ImageNet dataset and calculate both the mean and variance of the features associated with the all its datapoints and also the corresponding statistics for specific class datapoints.
\begin{align}
\label{eq:mean-prompts}
    \bm{\mu}(X ; l,\theta_v:\bm{p}) = \frac{1}{N} \sum_{\mathrm{x} \in \mathcal{A}(X)}  \bm{f^{l,p}}_{\mathrm{x}} \quad ,
\end{align}
\begin{align}
\label{eq:sigma-prompts}
    \bm{\sigma^2}(X ;l,\theta_v:\bm{p}) = \frac{1}{N} \sum_{\mathrm{x} \in \mathcal{A}(X)} \bigg(\bm{f^{l,p}}_{\mathrm{x}} - \bm{\mu}(X ;l, \theta_v:\bm{p})\bigg)^2,
\end{align}
where $\mathcal{A}(X)$ denotes the set of $N$ augmentations corresponding to the test sample $X$ and $\bm{f^{l,p}}_{\mathrm{x}}$ denotes the visual features corresponding to the augmentation $x$ of $X$ at the layer $l$ of the visual encoder when the prompt is $\bm{p}$. $\theta_v$ is denotes the parameters of the visual encoder. By overloading notation, we can define the statistics of the source dataset as -
\begin{align}
\label{eq:source-statistics}
 \bm{\hat{\mu}}_{l} =  \bm{\mu}(\mathcal{S};l, {\theta}_v)  \quad  \text{and}\quad       \bm{\hat{\sigma}^2}_{l} =  \bm{\sigma}^2(\mathcal{S};l, {\theta}_v) \quad , 
\end{align}
where $\mathcal{S}$ denotes the source dataset.
\newline
We further define the class statistics as -
\begin{align}
\label{eq:source-statistics}
 \bm{\hat{\mu}}_{l,c} =  \bm{\mu}(\mathcal{S}_c;l, {\theta}_v)  \quad  \text{and}\quad       \bm{\hat{\sigma}^2}_{l,c} =  \bm{\sigma}^2(\mathcal{S}_c;l, {\theta}_v) \quad , 
\end{align}
where $\mathcal{S}_c$ denotes the set of datapoints of the class $c$ of the source dataset.

Finally, we define the coarse alignment and finegrained alignment losses as - 
\begin{align}
\label{eq:coarse}
\begin{split}
    \mathcal{L}_{\text{coarse}} &= \frac{1}{L}\sum_{l=1}^{L} \bigg( \|  \bm{\mu}(X ; l,\theta_v:\bm{p}) - \bm{\hat{\mu}}_{l} \|_1  \\ &+  \| \bm{\sigma^2}(X ;l,\theta_v:\bm{p}) -   \bm{\hat{\sigma}^2}_{l}\|_1 \bigg).
    \end{split}
\end{align}
and
\begin{align}
\label{eq:fine}
\begin{split}
    \mathcal{L}_{\text{fine}} &= \frac{1}{L}\sum_{l=1}^{L} \bigg( \|  \bm{\mu}(X ; l,\theta_v:\bm{p}) - \bm{\hat{\mu}}_{l,c} \|_1  \\ &+  \| \bm{\sigma^2}(X ;l,\theta_v:\bm{p}) -   \bm{\hat{\sigma}^2}_{l,c}\|_1 \bigg).
    \end{split}
\end{align}
respectively. In \eqref{eq:fine} we assume that $X$ is an actively labelled sample which belongs to class $c$.

Our composite loss function is of the form 
\begin{align}
\label{eq:comp-loss}
\mathcal{L}=\mathcal{L}_{\textit{entropy}}+\alpha\mathcal{L}_{\textit{cross-entropy}}+\beta\mathcal{L}_{\textit{coarse}} + \gamma\mathcal{L}_{\textit{fine}}
\end{align}
%$$\mathcal{L}=\mathcal{L}_{\textit{entropy}}+\alpha\mathcal{L}_{\textit{cross-entropy}}+\beta\mathcal{L}_{\textit{align}}$$
Where $\mathcal{L}_{\textit{entropy}}$ is expressed in \eqref{eq:ent-min}, $\mathcal{L}_{\textit{cross-entropy}}$ is the standard supervised cross-entropy loss applied on actively labelled samples, $\mathcal{L}_{\textit{coarse}}$ is expressed in \eqref{eq:coarse} and $\mathcal{L}_{\textit{fine}}$ is given in \eqref{eq:fine}. It is also important to note that except for domain generalisation tasks, we make $\beta = 0$ and $\gamma = 0$; that is, we do not use explicit domain alignment.  
\SetKwComment{Comment}{\hspace{0.2em} \# }{}
\SetCommentSty{textnormal}

\begin{algorithm}[htbp]
\caption{Class Balanced Eviction from Buffer($\mathcal{D}_l$)}
\label{alg:Algorithm 2}

\KwIn{Unlabeled Dataset $\mathcal{D}_u$, Oracle$(\cdot)$}
\textbf{Initialize:} $N_{\text{queried}} = 0$, $\hat{\mu}_0 = 0$, $\hat{\sigma}_0 = 0$

\For{$t = 1, 2, 3, \ldots, |\mathcal{D}_u|$}{
    $\hat{\mu}_t, \hat{\sigma}_t \leftarrow \hat{\mu}_{t-1}, \hat{\sigma}_{t-1}$

    $\tau_t \leftarrow \text{\small{DynamicThresholdSelection}}(t, N_{\text{queried}}, \hat{\mu}_t, \hat{\sigma}_t)$

    \If{$H(\hat{x}_t) > \tau_t$}{
        $y_t \leftarrow \text{Oracle}(x_t)$

        $N_{\text{queried}} \leftarrow N_{\text{queried}} + 1$

        \If{$\mathcal{D}_l$ is full}{
            \Comment{Select class $m$ with most samples in $\mathcal{D}_l$}
            $m \leftarrow \arg\max_{k \in \{1, \ldots, K\}} |c_k|$

            \eIf{$|m| = 1$}{
                \Comment{Select sample with lowest CE loss in class $c_m$}
                $j \leftarrow \arg\min_{i \in \{1, \ldots, J\}} L_{\text{CE}}(x_i, y_i) \mid y_i = c_m$
                
                Remove $(x_j, y_j)$ from $\mathcal{D}_l$
            }{
                \Comment{Multiple max-size classes. Pick one with lowest avg CE loss}
                $l \leftarrow \arg\min_{k \in m} \bar{L}_{\text{CE}}(x_i, y_i) \mid y_i = c_k$

                \Comment{Select sample with lowest CE loss in class $c_l$}
                $j \leftarrow \arg\min_{i \in \{1, \ldots, J\}} L_{\text{CE}}(x_i, y_i) \mid y_i = c_l$

                Remove $(x_j, y_j)$ from $\mathcal{D}_l$
            }
        }

        Add $(x_t, y_t)$ to $\mathcal{D}_l$
    }
}
\end{algorithm}

%% file: ICCV2025-Author-Kit-Feb/sec/3_theoreticaljustification.tex
\section{Theoretical Justification}

The simplicity of our method makes it amenable to theoretical analysis which helps us formalize our intuition. While this theoretical machinery provide the foundational scaffolding for our approach, we emphasize that real-world implementation necessitates pragmatic adaptations to address computational constraints and operational realities. The experimental realization thus embodies a rational approximation of theoretical ideals – maintaining philosophical alignment while accommodating operational constraints through empirical validation. For conciseness, we present a succinct overview of our methodological framework in this section, with comprehensive technical elaboration contained in the supplementary materials.

\begin{proposition}[Class-Balanced Buffer Equilibrium]\label{prop:balance_revised}
Under the class-balanced replacement policy (Algorithm~\ref{alg:Algorithm 2}) with buffer capacity \(L\) and annotation budget \(B \gg L\), let
\[
f_{\mathrm{balance}}(\mathcal{D}_l^t) = \sum_{c=1}^K \left|\, |\mathcal{D}_l^{c,t}| - \frac{L}{K} \right|
\]
denote the class imbalance measure at time \(t\) of the buffer ($\mathcal{D}_l$). Then, with probability tending to one as \(B\to\infty\), the buffer achieves asymptotic class balance in the sense that
\[
\limsup_{t\to\infty}\left|\, |\mathcal{D}_l^{c,t}| - \frac{L}{K} \right| \le 1,\quad \forall\, c \in \{1,2,\dots,K\}.
\]
Equivalently, the probability of failure to reach balance equilibrium vanishes
\[
\lim_{B\to\infty} \mathbb{P}\!\left(f_{\mathrm{balance}}(\mathcal{D}_l^t) > 2\right) = 0.
\]
\end{proposition}
\begin{proof}
Here we just provide an informal proof outline with the complete proof with the technical details is being deferred to section \ref{pf:class_balance} of the appendix.  We begin our analysis after the buffer is full for the first time and the eviction policy comes into play. The critical observation is that the value of the imbalance function $f$ cannot increase from that point onward, except when it reaches zero. Further, even at zero it can only undergo slight perturbation and then return to the equilibrium state. After that, we show that reaching equilibrium is a high probability event which becomes a certainty in the asymptotic regime using lemma \ref{limit_lemma}.
\end{proof}

\begin{proposition}[Adaptive Query Convergence]\label{prop:query}
Let 
\[
R_N = \frac{1}{N}\sum_{i=1}^N \mathds{1}\Bigl\{X_i > \hat{\mu}_i + z_i \hat{\sigma}_i\Bigr\}
\]
be the empirical query ratio based on adaptive thresholds 
$
T_i = \hat{\mu}_i + z_i\hat{\sigma}_i,
$ where $\mathds{1}$ is the indicator function and where \(\hat{\mu}_i\) and \(\hat{\sigma}_i\) are online estimates of the mean and standard deviation computed from the samples \(\{X_1,\dots,X_i\}\) which are the cross-entropy values of images in the test stream which are assumed to be i.i.d.\ with
\[
X_i \sim \mathcal{N}(\mu,\sigma^2).
\]
For a target query ratio \(\alpha \in (0,1)\) (in our experiments, \(\alpha=0.05\)) and any \(\epsilon > 0\), we have
\[
\lim_{N \to \infty} \mathbb{P}\Bigl(|R_N - \alpha| \geq \epsilon\Bigr) = 0.
\]
The threshold scaling factors \(z_i\) are chosen adaptively via the switching rule:
\[
z_i = \begin{cases}
z_{\mathrm{high}}, & \text{if } R_{i-1} \geq \alpha, \\
z_{\mathrm{selection}}, & \text{if } R_{i-1} < \alpha,
\end{cases}
\]
so that the ideal per-image query probability is set to \(\alpha\).
\end{proposition}
\begin{proof}
  Here we provide an informal outline of the proof with the technical details being deferred to section \ref{pf:query_adaptive}. Compared to the previous proof this involves a more sophisticated probabilistic analysis using martingales. In the first part, we bound the estimation error of the threshold by bounding that of the mean and variance using standard concentration inequalities. Then we bound the estimation error of the probability of selection of a sample using the mean value theorem and Azuma-Hoeffding inequality. Using contradiction we prove that the overquerying and the consequent mode switch is asymptotically small. Consequently, it follows that the query rate converges in probability to the ideal query rate. 
\end{proof}

%% file: ICCV2025-Author-Kit-Feb/sec/4_experiment.tex
\section{Experiment}

Due to space constraints, we are not able to include all details in this section. More details regarding the experimental setup and ablation studies are given in sections \ref{sec:experimental_setup} and \ref{sec:ablation_studies} of the appendix respectively. Our framework is designed to be model-agnostic; however, we demonstrate it using Vision-Language Models because the widespread use of VLMs in various applications ensures that our method is tested under practical, real-world conditions, thereby underscoring its generality and effectiveness beyond any domain-specific constraints.

\begin{table*}[!hbt]
\caption{Results on cross-dataset transfer - the Top-1 accuracy is reported}
\centering
 \begin{tabular}{||c |c c c c c c c c c c c||} 
 \hline
 % 3rd to mid between 8th and 9th
  & \begin{turn}{90}Caltech101 \end{turn}& \begin{turn}{90} OxfordPets \end{turn} & \begin{turn}{90}Flowers\end{turn} & \begin{turn}{90}Food101\end{turn} & \begin{turn}{90}FGVC-Aircraft\end{turn} & \begin{turn}{90}SUN397\end{turn}& \begin{turn}{90}DTD\end{turn} & \begin{turn}{90}StanfordCars\end{turn} & \begin{turn}{90}EUROSAT\end{turn}& \begin{turn}{90}UCF101\end{turn} & \begin{turn}{90}Average\end{turn} \\ [0.5ex] 
 \hline\hline
 CLIP & 93.35 & 88.25 &  67.44 & 83.65 & 23.67 & 62.59 & 44.27 & 65.48 & 42.01 & 65.13 & 63.58 \\ 
 CLIP+TPT & 94.16 & 87.79 & 68.98 & 84.67 & 24.78  & 65.50 & 47.75 & 66.87 & 42.44 & 68.04 & 65.10  \\
 CoOp & 93.70 & 89.14 &  68.71 & 85.30 & 18.47 & 64.15 & 41.92 & 64.51 & 46.39 &  66.55 & 63.88 \\
 CoCoOp & 93.79 & 90.46 & 70.85 & 83.97 & 22.29 & 66.89 & 45.45 & 64.90 & 39.23 &  68.44 & 64.63\\
 ProDA & 86.70 & 88.20 & 71.50 & 80.80 & 22.20 & - & 40.90 & 60.10 & 48.50 & - & 65.62 \\
 MaPLe & 93.53 & 90.49 & 72.23 & 86.20 & 24.74 & 67.01 & 46.49 & 65.57 & 48.06 & 68.69 & 66.30  \\
 MaPLe+TPT & 93.59 & 90.72 & 72.37 & 86.64 & 24.70 & 67.54 & 45.87 & 66.50 & 47.80 & 69.19 & 66.50 \\
 PromptAlign & 94.01 & \textbf{90.76} & 72.39 & 86.65 & 24.80 & 67.54 & 47.24 & \textbf{68.50} & 47.86 & 69.47 & 66.92 \\[1ex] 
 \hline\hline
 ours & \textbf{94.49} &90.68 & \textbf{72.49}&\textbf{86.69}&\textbf{24.92}
&\textbf{67.59} &\textbf{48.02}& 67.84&\textbf{50.74} &\textbf{70.49}&\textbf{67.40} \\ [0.5ex]
 \hline
 \end{tabular}
 \label{Table 2}
\end{table*}

%\newline
\begin{table*}[h!]
\caption{Results on adaptation datasets - the zero-shot Top-1 accuracy is reported}
\centering
 \begin{tabular}{||c | c |c |c |c |c||} 
 \hline
 &Imagenet-V2 & ImageNet-Sketch & ImageNet-A & Imagenet-R & OOD Avg.\\\hline
  CLIP&60.86 & 46.09 & 47.87 & 73.98 & 57.20 \\
  CLIP+TPT & 64.35 & 47.94 & 54.77 & 77.06 & 60.81 \\
CoOP & 64.20 & 47.99 & 49.71 & 75.21 & 59.28  \\
  Co-CoOp & 64.07 & 48.75 & 50.63 & 76.18 & 59.91\\
  Co-Coop+TPT & 64.85 & 48.27 & 58.47 & 78.65 & 62.61\\
  PromptAlign& 65.29  & \textbf{50.23} & 59.15 & 79.02 &63.42\\[1ex]
\hline\hline
  ours &\textbf{65.60} & 50.14 & \textbf{59.31} & \textbf{79.51} & \textbf{63.64}\\[0.5ex]
 \hline
 \end{tabular}
 \label{Table 3}
\end{table*}

\subsection{Evaluation Protocol}
\label{eval_protocol}
In the standard TPT setting, with the MEM (Section \ref{tpt}) framework, there is an update on the unsupervised loss before the evaluation. In our setting, we also update on the unsupervised loss before evaluation, but we can't use a sample to calculate the supervised loss before evaluation since it doesn't make sense to evaluate a sample after already knowing its true label. In our evaluation protocol, we update on the unsupervised loss before evaluating on the sample, but actively label it only after the evaluation. After active labelling, the sample is put in the buffer and can be used in calculating the $\mathcal{L}_{\textit{cross-entropy}}$ in \eqref{eq:comp-loss} for the successive time steps. This ensures fairness in evaluation since we are using the true label to update the model only after a sample has been evaluated. This also means that we can evaluate prior arts with their usual evaluation scheme.
\subsection{Results}
The results of fine-grained classification are given in Table \ref{Table 2} while those of domain generalisation are given in Table \ref{Table 3}. It can be seen that in nearly all of the datasets in fine-grained classification we are exceeding all other baselines in terms of performance. Our average performance exceeds the next best by 0.48\% in transfer learning datasets, and in domain generalisation, our average performance exceeds by $0.22\%$. While in ImageNet R and ImageNet A we show better performance than the SOTA, in ImageNet Sketch our performance is marginally worse. The reason our performance on the adaptation datasets is overall not that impressive is because due to memory constraints our buffer size could only be very small (75 for some datasets). Still, for the sake of completeness we provided these results. Our average inference latency per sample, when the buffer size is at its maximum, is around 0.63s, which is about 50\% more than that of our baseline PromptAlign \citep{abdul2024align}, whose latency was found to be 0.41s when averaged over all 14 datasets. However, these figures must be contextualized in comparison to those from ATTA\citep{gui2024active}, where they observed up to almost \textbf{10x} increase in latency compared to their baselines.

\begin{figure*}[h]  
  \centering
    \includegraphics[width=1\linewidth]{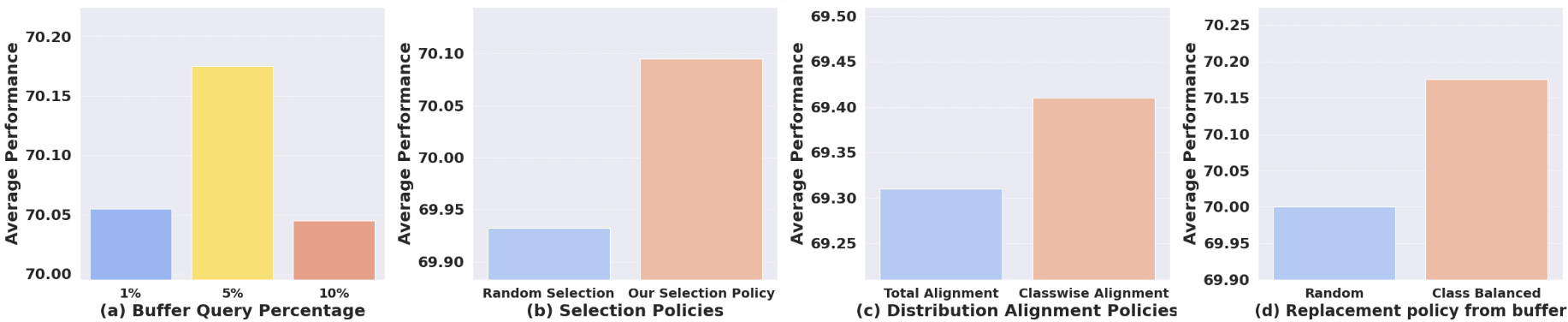}
    \caption{(a). Results of varying active samples query percentage, $5\%$ active query percentage was found to be optimum (b).Selecting random images vs our selection policy (c). Class balanced vs random replacement from $\mathcal{D}_l$ (d).  Total (Coarse-grained) alignment vs Classwise (Fine-grained) Alignment}
    \label{fig:Figure 2}
 \end{figure*}
 
\subsection{Ablation studies}
Due to space constraints we have shifted the detailed ablations section to section \ref{sec:ablation_studies} in the appendix. Readers are requested to refer to it for plots and detailed descriptions.
The ablation studies explore several design choices affecting model performance and robustness on challenging datasets such as Caltech101, DTD, Stanford Cars, and UCF101.\newline \textbf{Active samples queried percentage:} First, varying the active sample query percentage revealed that although increasing the number of actively queried samples generally improves performance by enhancing model robustness, there exists an optimal annotation budget. Experiments with 1\%, 5\%, and 10\% budgets showed that around 5\% yields the best performance, as evident in Figure \ref{fig:Figure 2} a. This optimum is likely due to a limited buffer capacity, where higher budgets lead to premature replacement of samples before fully extracting their information.\newline
\textbf{Changing the loss coefficient:} Next, the impact of the loss coefficient was examined by adjusting the parameter $\alpha$ in the composite loss function. The results indicate that emphasizing the unsupervised loss with $\alpha = 1$ provides the best average performance which is represented in Figure \ref{fig:loss coefficient} of the appendix, suggesting a balanced contribution between the supervised and unsupervised components.\newline
\textbf{Buffer size:} The studies then evaluated the buffer size by testing values of 25, 50, 100, and 150. Since storing all actively queried samples would be impractical, the maximum buffer size was capped at 150. The experiments confirmed that a larger buffer size enhances average performance, as it retains more useful information as clear from Figure \ref{fig:buffer size} of the appendix. \newline
\textbf{Random selection for $\mathcal{D}_l$ vs our selection policy:} Furthermore, the proposed informative sample selection policy was compared with random selection to populate the buffer. Under a 5\% query rate and a 150-sample buffer, the selection policy outperformed random sampling which is clear from Figure \ref{fig:Figure 2} b, highlighting its effectiveness in capturing critical information.\newline
\textbf{Class balanced vs random replacement from $\mathcal{D}_l$:} Additional experiments compared class-balanced replacement with random replacement from the labeled dataset. The class balancing approach, implemented using a dedicated algorithm, was evaluated on the same four datasets, demonstrating its advantages in maintaining performance stability as clear from Figure \ref{fig:Figure 2} c.\newline
\textbf{Total vs Classwise alignment:} Finally, a comparison between total (coarse-grained) and Classwise (fine-grained alignment) (evaluated on ImageNet-A and ImageNet-R) showed that fine-grained alignment yields superior performance, underscoring its effectiveness in aligning distributions more accurately which is evident from Figure \ref{fig:Figure 2} d.

\section{Potential Applications}
\label{practical-utility}
For the sake of fair comparison with previous arts, we actively label our samples only after evaluating them. However, it is important to note that, when it comes to practical application, we can also reverse the order. That is, we can ask the expert for their advice and take that as the ground truth. This is especially true because we are not restricted by our methodology to postpone the active labelling decision but instead do so in real-time, unlike \citep{gui2024active} where they collect the unlabelled samples in a buffer and select which ones to query later. Our single test sample in a time-step assumption makes our setting even more practical.
\newline\newline
Ideal scenarios for practical application are high-risk ones like autopilot systems and medical diagnosis, where the extra cost and latency in consulting an expert is justified by the potential avoidance of hazards. In autopilot systems, when the system is uncertain, it can hand over control to the pilot, who then demonstrates the correct way of handling that particular scenario. The system then stores the pilot's actions in memory and uses them to learn the correct way so that it does not need human intervention in a similar scenario in future. In medical diagnosis, whenever the system is uncertain, instead of making a wrong diagnosis, it sends the sample over to a medical practitioner who then provides his expert opinion.

%% file: ICCV2025-Author-Kit-Feb/sec/7_conclusion.tex
\section{Conclusion}
In this paper, we tackle the problem of Active Learning of VLM prompts in a Test Time setting. Building on prior work which demonstrated the relevance of Active Learning in a Test Time setting, we extend that to Prompt Learning in VLMs. We further demonstrate the relevance by showing that marginal entropy minimisation on uncertain samples reinforces errors in the model. This makes knowing the true label even more relevant. Ours is the first work to deal with the problem of Active Test Time Optimisation with a single test sample at a time. We use a dynamically adjusted threshold for entropy based on which we select the samples which will be chosen for active labelling and we do so in such a manner that our annotation budget is not exceeded or exhausted too early into the data streaming process. Our method uses a replacement policy which prioritises class balancing and informativeness of samples to make intelligent use of the limited-size buffer. We propose a fair evaluation protocol which shows that Active Learning is indeed an effective strategy in the Test Time Prompt Learning of VLMs.

%% file: ICCV2025-Author-Kit-Feb/sec/X_suppl.tex
\clearpage
\setcounter{page}{1}
\maketitlesupplementary

\section{Preliminaries}
\label{sec:preliminaries}
\subsection{Active Learning}
In active learning, we have an unlabelled dataset $\mathcal{D}_u$. The typical setting is that of multi-class classification, having $K$ classes. The training happens in an iterative way where, in each iteration, the model selects some samples from $\mathcal{D}_u$, which it then passes on to the oracle to get the correct annotations. At each iteration, labelled samples are added to the labelled dataset $\mathcal{D}_l$. The labelled dataset is used to train the model before the next query phase, after which the size of $\mathcal{D}_l$ further increases. There is typically also a budget constraint that the model has to maintain. That is, the total number of queried samples cannot exceed a budget of $B$.
\subsection{Vision Language Models and Prompt Learning}
In Vision Language Models (VLMs) like CLIP \cite{radford2021learning}, training is performed by pairing an image and the corresponding caption. There are two encoders - an image encoder $E_v$ and a text encoder $E_t$. The image and text are passed through the encoders where the image encoder is either ViT \cite{dosovitskiy2020image} or ResNet \cite{he2016deep} architectures while the text encoder uses the Transformer \cite{vaswani2017attention,devlin2018bert} architecture. 

Once the image $x$ and text input $t_i$ (which is typically a classname out of the $K$ classnames) are mapped into the embedding space, which can be represented as 
$$e_v = E_v(x)\,\,\,\,\,\,\,\,\,\,\,\,\,\,e_t^i = E_t(\{t_i;p\})$$
where $p$ is a handcrafted prompt prepended to the text input.
Next, cosine similarity is computed between the visual embedding $e_v$ and textual embedding $e_t$. The objective of contrastive training is to reduce the distance between correct pairings and increase that between incorrect ones. That also achieves the objective of ensuring that the following prediction probability distribution
$$P(c=i|x) = \frac{exp(cos(e_v, e_t^i)/\tau)}{\sum_{j=1}^Kexp(cos(e_v, e_t^j)/\tau)}$$
concentrates more of the density towards the correct class. Here $cos$ denotes the cosine similarity.
Prompt Learning essentially makes the optimisation of a pre-trained VLM easier by only optimising a small parameter group known as prompts which are auxiliary to the original VLM. Instead of handcrafted prompts, which may require a lot of manual tuning, prompts are used as learnable vectors. The prompts are concatenated to the text input which is then passed through the text encoder to be mapped into the embedding space. There also exists a variant of prompting known as Multimodal Prompt Learning \cite{khattak2023maple} where learnable vectors are attached to both visual and textual tokens. In this paper, we use multimodal prompt learning.
\subsection{Test Time Prompt Tuning}\label{tpt}
In TPT \cite{shu2022test}, they have the same prompt tuning setting except that the prompt tuning has to be done on the fly on a single test sample $x_t$ at a time $t$. They then use a set of augmentation functions $\mathcal{A}$ to make $n$ augmentations $\Tilde{x}_t= \mathcal{A}(x_t)$. $\Tilde{x}_t$ is then passed through $E_v$ to get the logits corresponding probability distribution over classes for each augmentation. The entropy for each of the $n$ logits is found, and only the top $\rho$ logits with the least entropy are retained. The logits are averaged and the entropy of the average logit is computed. The marginal entropy of this average logit is the training objective at each time step $t$. This is called Marginal Entropy Minimisation (MEM) as we try to minimise this objective.
\begin{align}
\label{eq:ent-min}
    % \bm{p}^{\ast} 
    \mathcal{L}_{\text{entropy}}= -\sum_{i=1}^{C}\Tilde{p}_{\bm p} (y_i|X_{\text{test}}) \log \Tilde{p}_{\bm p}(y_i|X_{\text{test}})
    % \text{where } \quad &\Tilde{p}_{\bm p}(y_i|X_{\text{test}}) = \frac{1}{N}\sum_{i=1}^{N}p_{\bm p}(y_i|\mathcal{A}_i(X_{\text{test}})).
\end{align}

where $\bm p$ are the learnable prompts and $\Tilde{p}_{\bm p}(y_i|X_{\text{test}})$ represents the mean of vector class probabilities produced by the model across the different augmented views preserved after the confidence selection filter. 
\section{Related Work}\label{app:related work}
\subsection{Active Learning}
Active Learning promotes label efficiency by imposing a label budget. It can be used in a variety of settings to gain knowledge about some aspect that is not already known and the model is uncertain about via an oracle. The queried samples are then sent to an oracle who returns the true label. Other than choosing on the basis of uncertainty \citep{lewis1994heterogeneous,yang2016active,roth2006margin,holub2008entropy}, in latest works, the model also tries to maintain diversity \citep{parvaneh2022active,sener2017active} in its choice of samples to query. There is typically a dynamic buffer that is maintained by the Active Learning algorithm wherein the annotated samples are put. The buffer is enlarged as more samples which are both diverse and informative are added to it. There have been incorporations of Active Learning into Domain Adaptation as well \citep{prabhu2021active}. In \citep{wang2022activesourcefreedomain, kothandaraman2022saladsourcefreeactivelabelagnostic} incorporated Active Learning into Source Free Domain Adaptation which, while not directly dependent on the sourced data, are also not suited for continuous data streams, unlike our TTA setting. In \citep{pmlr-v202-saran23a}, they take up the task of actively labelling samples in a streaming setting. However, their work significantly differs from ours as they don't continuously adapt their parameters as the data stream progresses. Instead, they reinitialise their parameters to the original parameters after each new data point is acquired. A contemporary work that is more closely related to ours is \citep{gui2024active}, where they provide some foundational theoretical work on Active Test Time Adaptation. However, their work is also significantly different from ours because they assume a batch setting wherein at each timestep, a minibatch comes for inference while we only assume a single sample. They also do multiple gradient updates in each time step while we do only one. Their labelling is also not done in real-time but effectively postponed by placing the unlabelled samples in a buffer from which they have to be selected for active labelling later. This may not always be possible due to privacy and storage concerns where it may not be feasible to postpone the decision of sending a sample to an oracle and instead retaining it. On the other hand, we make the active labelling decision in real-time based on our dynamically adjusted threshold.
\subsection{Test Time Adaptation}
Adaptation of pre-trained models is necessary so that they can be used optimally for the given task at hand. Fully Test Time Adaptation \citep{nado2020evaluating,schneider2020improving,sun2023vpa,liang2023comprehensive} is the highly realistic and practical setting wherein a pre-trained model has to optimise and adapt its parameters to a situation that  it faces during inference, that is, amidst real-time use. A prominent example includes that of a self-driving car where the car has to adapt to unforeseen conditions while it is being used. A popular way to achieve fully test time adaptation has been to update the statistics of the Batch Normalisation layer during inference. In TENT \citep{wang2020tent}, the Batch Normalisation parameters are updated using a self-entropy objective. However, TENT makes the batched input assumption. In MEMO \citep{zhang2022memo}, they use the more general case of a single test input by taking multiple augmentations of a single image. TPT \citep{shu2022test} essentially extends the MEMO philosophy to prompt tuning to make VLMs adapt at test-time by updating prompts. Especially relevant to this paper is the newly introduced paradigm of Active Test Time Adaptation \cite{gui2024active}, where the model has the option of querying a few samples during test time adaptation. It was found that at a some latency cost as compared to FTTA methods, the ATTA framework provides much superior results, and thus, its use-case may not completely overlap with that of FTTA.
\subsection{Prompt Learning in VLMs}
Prompt Learning has been proposed as a method of fine-tuning VLMs in compute-constrained scenarios due to its parameter efficiency. In CoOp \citep{zhou2022learning} and CoCoOp \citep{zhou2022conditional}, the prompts are appended to the textual tokens and help provide context to the input of the VLM instead of finetuning the entire VLM model. Learning good prompts has been shown to dramatically improve the performance of the CLIP \citep{radford2021learning} model. Prompt Learning has been extended as a transfer learning and adaptation method in various ways and settings. Of special interest to us is the Test Time setting \citep{shu2022test}. The Test Time scenario is highly practical given that foundation models like VLMs are becoming more mainstream and adapting them at test time by optimising a small parameter group like prompts is likely to be the way forward. More recently, in \citep{bang2024active} they introduce Active Learning to Vision Language models where it is noted that diversity in the form of class-balancing is important for non-trivial gains in VLM performance via Active Learning. However, our method is significantly different from theirs because it is in a test time scenario, whereas theirs was in a supervised learning setting.
\section{Experimental Setup}
\label{sec:experimental_setup}
\textbf{Datasets.}
In domain generalization, we evaluate on four out-of-distribution (OOD) variants of ImageNet \cite{deng2009imagenet}; ImageNet-Sketch \cite{wang2019learning},ImageNet-A \cite{hendrycks2021natural}, ImageNet-V2 \cite{recht2019imagenet} and ImageNet-R \cite{hendrycks2021many}. For cross-dataset transfer, we try on 10 image classification datasets which cover a wide variety of visual recognition tasks. Among these Caltech101 \cite{fei2004learning}; five datasets which are fine-grained StanfordCars \cite{krause20133d},Flowers102 \cite{nilsback2008automated},OxfordPets \cite{parkhi2012cats},Food101 \cite{bossard2014food} and FGVC-Aircraft \cite{maji2013fine}, which contain images of transportation, flowers and animals; and four datasets of textures, satellite imagery, scenes and human actions which are DTD \cite{cimpoi2014describing}, EUROSAT \cite{helber2019eurosat}, SUN397 \cite{sun2020test} and UCF101 \cite{soomro2012ucf101} respectively.\newline

\textbf{Implementation Details.} Following PromptAlign \cite{abdul2024align}, using a single test sample we optimize the prompts on both the text and vision branches. Our models were implemented on a single NVIDIA A40 48GB GPU using the PyTorch framework. Refer to section \ref{tpt}, we take $n = 63$ and $\rho = 10\%$. Then we compute the token distribution alignment loss between the tokens of all the 64 images(\eqref{eq:fine}). A learning rate of $5e^{-4}$ was used for the fine-grained datasets Flowers102, OxfordPets, Food101, SUN397, FGVCAircraft, and EuroSAT and a learning rate of 0.004 for the rest of the datasets. We use an annotation budget of $5\%$ and a buffer size of $150$ for all datasets except Imagenet-v2 and Imagenet-Sketch where we use a buffer size of 75 due to memory constraints. The static threshold $\tau$ for selecting samples to be queried for labeling was fixed at 2, till $\tilde{t}=30$ time steps. 
Refer to \eqref{eq:comp-loss}. All the results in Table \ref{Table 2} are produced by taking $\alpha=1$, $\beta = 0$ and $\gamma = 0$. On the other hand, in Table \ref{Table 3}, the values of $\alpha$ are 1 in Imagenet-R and Imagenet-V2, and 0.15 and 0.5 in Imagenet-A and Imagenet-Sketch respectively, $\beta = 1$ and $\gamma = \alpha$ for all. We hypothesize that the lower value of $\alpha$ suited for these datasets is due to a greater number of outlier samples present in them.

\textbf{Baselines.} We evaluate our method with existing few-shot prompt learning methods for adapting CLIP including CoOp \cite{zhou2022learning} and CoCoOp \cite{zhou2022conditional}, TPT \cite{shu2022test} and PromptAlign \cite{abdul2024align} method. MaPLe \cite{khattak2023maple} is a multi-modal prompt learning baseline, which adapts CLIP by learning deep prompts on both the text and vision branches. TPT is a test-time prompt tuning method that tunes the prompt at test time per input sample, which achieved strong performance in prompt learning when combined with CoOp. It is important to note that whenever we have appended +TPT to a method, it means that the corresponding supervised counterpart has been taken and executed with TPT loss.
\section{Ablation studies}
\label{sec:ablation_studies}
\subsection{Active samples queried percentage:}
\begin{figure}[h]
  
  \centering
    \includegraphics[width=.9\linewidth]{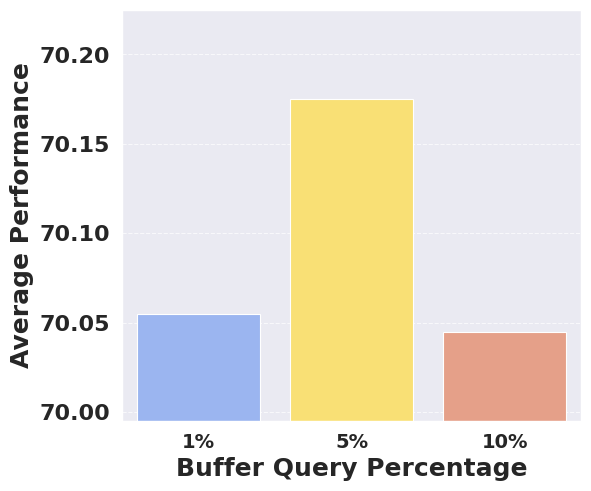}
    \caption{Varying active samples query percentage, 5\% active query percentage was found to be optimum}
    \label{fig:Figure 3a}
 \end{figure}

Increasing the number of samples actively queried increases the model's robustness and hence helps improve its performance, especially for more challenging datasets. But that comes at a cost of the annotation budget, since the annotation budget is quite limited. We tried our experiments for different annotation budgets of 1\%,5\%,10\%. The results of using different annotation budgets are given in Figure \ref{fig:Figure 3a} which were obtained by taking the average performance over Caltech101, DTD, Stanford Cars and UCF101 datasets. The results show that increasing the annotation budget does not necessarily increase gains - there is an optimum annotation budget, which we found to be around 5\%. We hypothesise that this is because of our limited buffer capacity which implies that with greater annotation budget, samples have to be replaced more often before all the information has been extracted from them. In any case, the performance is robust to change in query percentage as well - perhaps pointing to the fact that most of the information is contained in the top 1$\%$ (by entropy) of samples. 
\subsection{Changing the loss coefficient in loss function}
\begin{figure}[h]
  \centering
    \includegraphics[width=.9\linewidth]{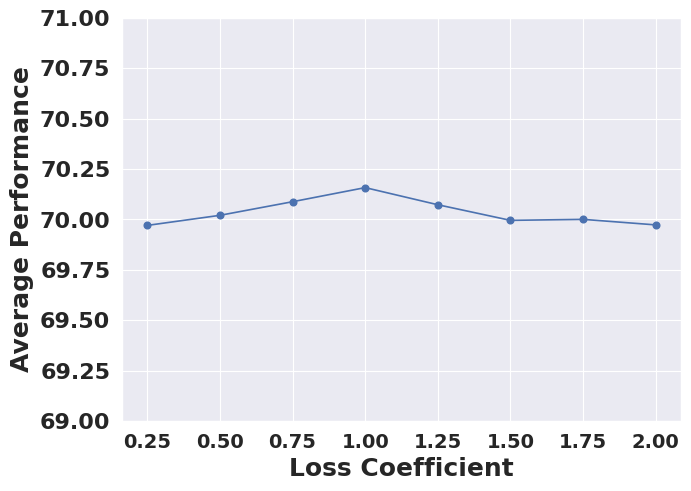}
    \caption{Changing the loss coefficient in loss function,  the optimum value was found to be 1}
    \label{fig:loss coefficient}
 
\end{figure}

When we increase the value of $\alpha$ in \eqref{eq:comp-loss}, it means giving more importance to the unsupervised loss so we change the value of $\alpha$ in this range, the results are given in Figure \ref{fig:loss coefficient} which were obtained by taking the average performance over Caltech101, DTD, Stanford Cars and UCF101 datasets. The results show that on average, the value $\alpha = 1$ is the optimum value.

\subsection{Changing the size of buffer for values 25,50,100,150:}
\begin{figure}[h]
  
  \centering
    \includegraphics[width=.9\linewidth]{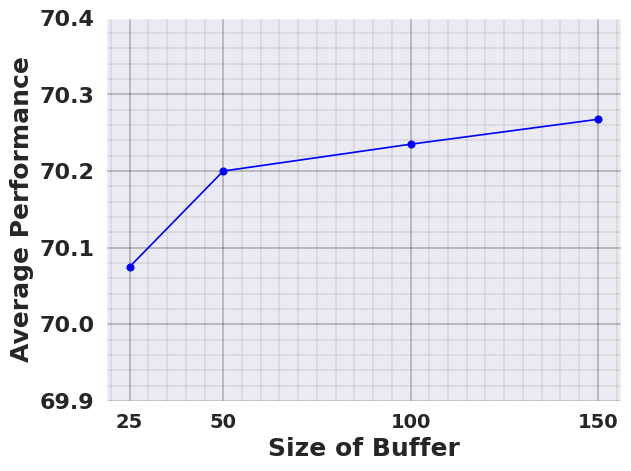}
    \caption{Varying the size of the buffer,  increasing the buffer size leads to increase in average performance }
    \label{fig:buffer size}
  \end{figure}

Now the size of the buffer was varied for the values 25,50,100 and 150 and the corresponding performance was checked. We can not arbitrarily have a very large buffer size i.e. storing all the actively queried samples since for datasets like Food101, 5\% of the total test samples is around 1500 images. Storing all those images in the buffer would increase the latency. So the maximum buffer size was set to 150. The results in Figure \ref{fig:buffer size} were obtained by changing the buffer size to the values mentioned above, while 5\% of the test samples were queried and taking the average performance for the datasets Caltech101, DTD, Stanford Cars, UCF101.

\subsection{Selecting random images for the buffer vs our selection policy}
\label{ablation:selection}
\begin{figure}[h]
  \centering
    \includegraphics[width=.9\linewidth]{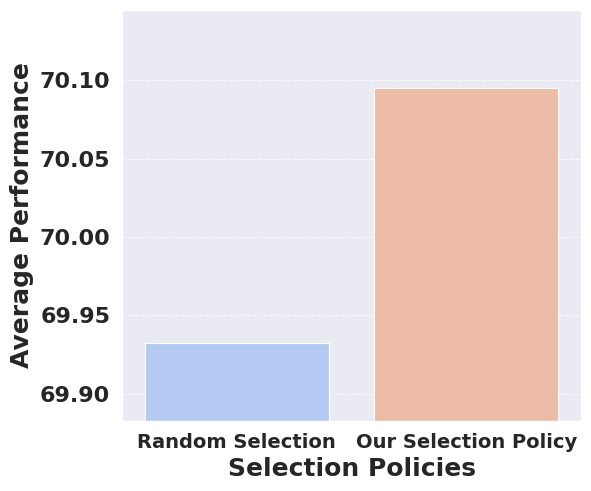}
    \caption{Selecting random images vs our selection policy, our image selection policy outperforms random image selection}
    \label{fig:Figure 4b}
 
\end{figure}

Our selection policy of selecting informative samples from the test images for the buffer was now compared to selecting random images from the buffer. This was performed with maximum buffer size of 150, query of 5\% test samples, and is the average of the performance in the Caltech101, DTD, Stanford Cars and UCF101 datasets. The results in Figure \ref{fig:Figure 4b} indicate that our selection policy is indeed shows better performance.

\subsection{Class balanced vs random replacement}
For class balancing we used Algorithm \ref{alg:Algorithm 2}, for a description of the class balancing method refer to section \ref{class balancing}.
\newline
\begin{figure}[h]
  \centering
  \includegraphics[width=0.45\textwidth]{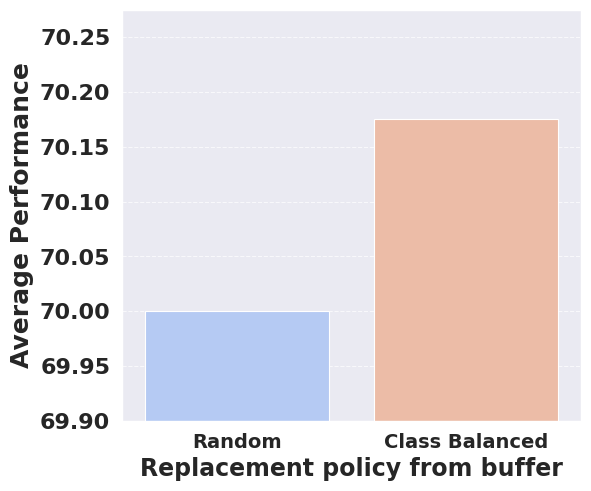}
  \caption{Class balanced vs random replacement from $\mathcal{D}_l$}
  \label{fig:Figure 7}
\end{figure}

For random replacement from $\mathcal{D}_l$, we randomly selected a class $c_k$ among all the classes such that $|c_k|\geq1$, i.e. a non empty class, and then randomly replaced a sample from it.\newline
It is evaluated on Caltech101, DTD, Stanford Cars, UCF101. And the results shown in Figure \ref{fig:Figure 7}
\subsection{Coarse Grained alignment vs Fine Grained Alignment}
\label{sec:dist-align}
\begin{figure}[h]
  \centering
  \includegraphics[width=0.45\textwidth]{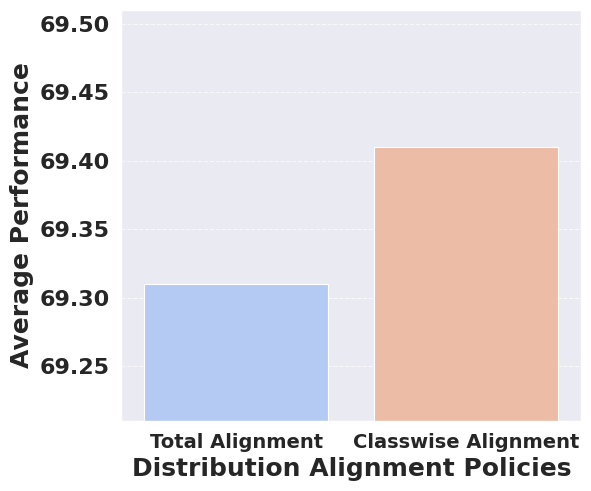}
  \caption{Total(Coarse-grained) alignment vs Classwise(Fine-grained) Alignment  }
  \label{fig:dist-align}
\end{figure}

In Figure \ref{fig:dist-align}, we show the difference in performance when we use coarse-grained alignment even on active samples. There is drop in performance which shows that fine-grained alignment is indeed effective. The evaluation has been done on ImageNet-A and ImageNet-R.
\section{Theoretical Discussion}
\subsection{Class Balance in Buffer}
\label{pf:class_balance}
Our policy is to first fill the buffer, not caring about the class of the image added. Once the buffer is full, then we remove images from the the class which has the maximum number of images for each newly queried image. If there are multiple maximum classes, then we select one of them which has the least average cross-entropy per image, and delete the image which has the least cross-entropy. Let the annotation budget be B, and the maximum capacity of the buffer be L, and the number of classes be m. 
\newline
Our goal is that by the end of the annotation budget, the number of images in each class should be close to $\frac{L}{m}$. Now we define a function which gives us a measure of how balanced the buffer is at the query of the $t^{th}$ image \begin{align}
    f_{balance}(\mathcal{D}_l^t)=\sum_{i=1}^m \biggl|n_i-\frac{L}{m}\biggl|
\end{align}
where $n_i$ is the number of images of class i present in the buffer. Hence it is clear that $f_{balance}(\mathcal{D}_l^t)=0$, if the buffer is completely balanced. 
Also when the buffer is full, then the maximum value of $f_{balance}(\mathcal{D}_l^t)$ is $2L\biggl(1-\frac{1}{m}\biggl)$, i.e. the case when only class has all L images, and all m-1 other classes are empty with 0 images. So when the buffer is full, we get the bound \begin{align}\label{inequality_balance}
    0\leq f_{balance}(\mathcal{D}_l^t)\leq 2L\biggl(1-\frac{1}{m}\biggl)
\end{align}
It is a trivial observation that for unbalanced buffer, the maximum class should have $n_{max}> \frac{L}{m}$, except the case of all classes being balanced and having $\frac{L}{m}$ images. So if there are multiple maximum classes, they all should have more than $\frac{L}{m}$ samples. Now $f_{balance}(\mathcal{D}_l^{t+1})$ remains the same value as $f_{balance}(\mathcal{D}_l^t)$ when the next queried image belongs to the maximum class, or any other class which has $\geq \frac{L}{m}$ samples, which is clear from the definition of $f_{balance}(\mathcal{D}_l^t)$, since the reduction of 1 due to the deletion of 1 image from max class, is neutralized by adding the queried image to the a class which has $\geq \frac{L}{m}$ samples. But if the image belongs to a class whose number of images $<\frac{L}{m}$, then in the next step, the value of $f_{balance}(\mathcal{D}_l^{t+1})$ reduces by 2 from the value of $f_{balance}(\mathcal{D}_l^{t})$, since 1 image is reduced from the max class, and the added image is added to a class which has 
$<\frac{L}{m}$ images, so the reduction is by 2. \begin{align}
\begin{array}{l} \label{cases}
f_{balance}(\mathcal{D}_l^{t+1}) = \\[6pt]
\begin{cases}
f_{balance}(\mathcal{D}_l^{t}) & \text{if } |\text{class}_{\text{queried}}| \ge \frac{L}{m},\\[4pt]
f_{balance}(\mathcal{D}_l^{t})-2  & \text{if } |\text{class}_{\text{queried}}| < \frac{L}{m}.
\end{cases}
\end{array}
\end{align}

where $|\text{class}_{\text{queried}}|$ indicates the number of images belonging to the class queried at that instant. So it is clear that $f_{balance}(\mathcal{D}_l^{t})$ is a non-increasing function, except for when $f_{balance}(\mathcal{D}_l^{t})$=0. 
\newline
\textbf{Balance equilibrium:}\newline
So if the initial value of the function is $f_{balance}(\mathcal{D}_l^{0})$, when the buffer is full for the first time, it is clear that $f_{balance}(\mathcal{D}_l^{t})$ will become 0 if there are $\frac{f_{balance}(\mathcal{D}_l^{0})}{2}$ times a query to a class which at that instant has $<\frac{L}{m}$ images. Also once $f_{balance}(\mathcal{D}_l^{t})$ becomes 0, then by definition all the classes are max classes, so if the next query is of the class having the lowest average cross entropy then, $f_{balance}(\mathcal{D}_l^{t+1})$ remains 0, else the class having the lowest average cross entropy gets 1 of its image removed, and the queried image is added to its respective class, so $f_{balance}(\mathcal{D}_l^{t+1})$, becomes 2 but from \ref{cases}, it is clear that it will not increase from that value, but can become 0 eventually. So it is observed that once the buffer gets balanced for the first time, it ends up in a cyclic equilibrium of $f_{balance}(\mathcal{D}_l^{t})$ being 0 and 2 only. \newline
\textbf{Probability of Failure to reach equilibrium:}\newline
We assume $B>>L$, following the practical case where the limit of the buffer is significantly small. Now we define $p_i$ to be the probability that the query is in a class which has $\geq \frac{L}{m}$ images at the $i^{th}$ instant. Now it is to be noted that $p_i<1$ strictly since $p_i=1$ indicates that the system has already reached the cyclic equilibrium so it can not fail. Now it will fail if during the whole process of querying B-L images the classes with $<\frac{L}{m}$ images at a particular instant are not queried at least $\frac{f_{balance}(\mathcal{D}_l^{0})}{2}$ times at those instants. Let the random variable X denote the number of times a class with $<\frac{L}{m}$ images at a particular instant are queried at those instants. So \begin{align*}
    \mathbb{P}(failure)&=\mathbb{P}(X=0)+\mathbb{P}(X=1)+...+\\ &\mathbb{P}\biggl(X=\frac{f_{balance}(\mathcal{D}_l^{0})}{2}-1\biggl)\\
    \end{align*}
    Now since $p_i$ can change in different scenarios of X having different values, so we denote $p_{i,x}$ as the sequence of $p_i$ values for a given value of X.\begin{align*}&\mathbb{P}(X=0)=\prod_{i=1}^{B-L}p_{i,0}, \\ &\mathbb{P}(X=1)=\binom{B-L}{1}\biggl(\prod_{i=1,i\neq j}^{B-L}p_{i,1}\biggl)(1-p_{j,1}),\\
    &...\\
    &\mathbb{P}\biggl(X=\frac{f_{balance}(\mathcal{D}_l^{0})}{2}-1\biggl)\\
    &=\binom{B-L}{\frac{f_{balance}(\mathcal{D}_l^{0})}{2}-1}\biggl(\prod_{i=1,i\neq j\,\,\forall j}^{B-L}p_{i,1}\biggl)\biggl(\prod_j(1-p_{j,1})\biggl),\\
    \end{align*} Now for each of these we can have an upper bound for each of these probabilities, let we define $v_0 = \max_{i \in I} p_{i,0}$ and for all $X>0$, we define $v_x = \max_{i \in I} \{ p_{i,x},\, 1-p_{i,x} \}$. Also it is clear that $v_0,v_1,...<1$ since all $p_{i,x}<1$. So we get upper bounds as \begin{align*}
        &\mathbb{P}(X=0)\leq v_0^{B-L}\\
        &\mathbb{P}(X=1)\leq \binom{B-L}{1}v_1^{B-L}\\
        &...\\
        &\mathbb{P}\biggl(X=\frac{f_{balance}(\mathcal{D}_l^{0})}{2}-1\biggl)\\ &\leq \binom{B-L}{\frac{f_{balance}(\mathcal{D}_l^{0})}{2}-1}v_{\frac{f_{balance}(\mathcal{D}_l^{0})}{2}-1}^{B-L}\\
    \end{align*} Now we define $v_{max} = \max_{x \in X} v_{x}$. We have $\frac{f_{balance}(\mathcal{D}_l^{0})}{2}-1 < L$ from \ref{inequality_balance}. So following these results we can say that \begin{align*}
        \mathbb{P}(failure)&\leq v_0^{B-L}+ \binom{B-L}{1}v_1^{B-L}+...\\
        &\binom{B-L}{\frac{f_{balance}(\mathcal{D}_l^{0})}{2}-1}v_{\frac{f_{balance}(\mathcal{D}_l^{0})}{2}-1}^{B-L}\\
        &<v_{max}^{B-L}+\binom{B}{1}v_{max}^{B-L}+...+\binom{B}{L}v_{max}^{B-L}
    \end{align*}
    Now we consider $S=v_{max}^{B-L}+\binom{B}{1}v_{max}^{B-L}+...+\binom{B}{L}v_{max}^{B-L}$, and consider the asymtotic behavior when B is very large. Since \(L\) is fixed and \(B>>L\)), then the sum
\[
\sum_{i=0}^{L} \binom{B}{i}
\]
grows only polynomially in \(B\) (approximately \(O(B^L)\). And so $B-L\approx B $

So we obtain the bound
\begin{align}
S = O\!\Big(B^L\;v_{max}^B\Big)
\end{align}

We use \ref{limit_lemma} and here since $v_{max}<1$ strictly, so $\frac{1}{v_{max}}>1$. So asymtotically S tends to 0. \newline So in conclusion $\mathbb{P}(failure)\approx 0$ asymtotically.
\begin{lemma}\label{limit_lemma}
Let \(L \in \mathbb{N}\) be fixed and \(c > 1\) be a constant. Then,
\[
\lim_{B \to \infty} \frac{B^L}{c^B} = 0.
\]
\end{lemma}

\begin{proof}
For \(B \ge 1\), define
\[
a_B = \frac{B^L}{c^B}.
\]
We show that for all sufficiently large \(B\), there exists a constant \(\rho\) with \(0 < \rho < 1\) such that
\[
a_{B+1} \le \rho\, a_B.
\]

First, observe that
\[
\frac{a_{B+1}}{a_B} = \frac{(B+1)^L}{B^L}\cdot \frac{1}{c} = \left(1+\frac{1}{B}\right)^L \frac{1}{c}.
\]
Since
\[
\lim_{B\to\infty} \left(1+\frac{1}{B}\right)^L = 1,
\]
for any \(\epsilon > 0\) there exists an integer \(B_0\) such that for all \(B \ge B_0\),
\[
\left(1+\frac{1}{B}\right)^L < 1+\epsilon.
\]
Choose \(\epsilon > 0\) sufficiently small so that
\[
\frac{1+\epsilon}{c} =: \rho < 1.
\]
Then, for all \(B \ge B_0\) we have
\[
\frac{a_{B+1}}{a_B} < \rho.
\]
This inequality implies that for every \(B \ge B_0\),
\[
a_{B+1} \le \rho\, a_B.
\]
By applying this inequality repeatedly, we obtain for any integer \(k \ge 0\):
\[
a_{B_0+k} \le \rho^k\, a_{B_0}.
\]
Since \(\rho < 1\), the term \(\rho^k\) converges to 0 as \(k \to \infty\). Hence,
\[
\lim_{B \to \infty} a_B = 0.
\]
\end{proof}
\subsection{Analysis of the Adaptive Query Policy}
\label{pf:query_adaptive}
\subsubsection{Policy description}
We design an adaptive policy for querying images whose cross-entropy losses are assumed to be independent $\mathcal{N}(\mu,\sigma^2)$ random variables. At each time, online estimates of the mean and variance are computed and used to set a threshold
\[
T_i = \hat{\mu}_i + z_i\,\hat{\sigma}_i,
\]
where the constant $z_i$ is chosen according to a switching rule. In particular, if the running query ratio up to time $i-1$ is below $7.5\%$, then $z_i=z_{0.05}$ so that the ideal per-image query probability is $0.05$, and if it is at least $7.5\%$, then $z_i=z_{0.025}$ so that the ideal probability is $0.025$. We rigorously prove that, under standard concentration assumptions, the overall query ratio converges to $5\%$, and we show by contradiction that the stricter ($2.5\%$) regime cannot persist for a nonzero fraction of time asymtotically.

\subsubsection{Analysis}

Assuming that $\{X_i\}_{i\ge 1}$ are i.i.d.\ random variables with
\[
X_i \sim \mathcal{N}(\mu,\sigma^2),
\]
and the online unbiased estimators are defined by
\[
\hat{\mu}_n = \frac{1}{n}\sum_{i=1}^n X_i,\quad \hat{\sigma}_n^2 = \frac{1}{n-1}\sum_{i=1}^n (X_i-\hat{\mu}_n)^2.
\]We assume that for $n \ge n_0$ there exist functions $\epsilon_1(n)$ and $\epsilon_2(n)$ satisfying
\[
\epsilon_1(n),\,\epsilon_2(n)\to 0 \quad \text{as } n\to\infty,
\]
since by the Strong Law of Large Numbers the sample mean
\[
\hat{\mu}_n = \frac{1}{n}\sum_{i=1}^n X_i
\]
converges almost surely to $\mu$. Also Hoeffding's inequality guarantees that for any $\delta>0$
\begin{align}\label{epsilon1}
\mathbb{P}\Bigl(|\hat{\mu}_n-\mu|\ge \epsilon_1(n)\Bigr) \le 2\exp\Bigl(-\frac{n\epsilon_1(n)^2}{2\sigma^2}\Bigr),
\end{align}
with $\epsilon_1(n)\to 0$ as $n\to\infty$. Similarly, under appropriate moment conditions, the sample standard deviation $\hat{\sigma}_n$ satisfies a concentration bound of the form
\begin{align}\label{epsilon2}
\mathbb{P}\Bigl(|\hat{\sigma}_n-\sigma|\ge \epsilon_2(n)\Bigr) \le 2\exp\Bigl(-c\,n\epsilon_2(n)^2\Bigr)
\end{align}
for some constant $c>0$. Hence, both $\epsilon_1(n)$ and $\epsilon_2(n)$ may be chosen to tend to zero as $n\to\infty$. And there exists a constant $\delta_1\in (0,1)$ such that
\begin{align}\label{both_epsilon_together}
\mathbb{P}\Bigl(|\hat{\mu}_n-\mu|\le \epsilon_1(n) \text{ and } |\hat{\sigma}_n-\sigma|\le \epsilon_2(n)\Bigr)\ge 1-\delta_1.
\end{align}
For each image $i$, we set a threshold
\[
T_i = \hat{\mu}_i + z_i\,\hat{\sigma}_i,
\]
with the ideal threshold being defined by
\[
T_i^* = \mu + z_i\,\sigma.
\]
On the event following \ref{both_epsilon_together} we have
\[
\mathcal{E}_i=\{|\hat{\mu}_i-\mu|\le \epsilon_1(i),\; |\hat{\sigma}_i-\sigma|\le \epsilon_2(i)\},
\]
and hence following triangle inequality we have
\[
|T_i-T_i^*|\le |\hat{\mu}_i-\mu|+|z_i|\,|\hat{\sigma}_i-\sigma|\le \epsilon_1(i)+|z_i|\epsilon_2(i) \triangleq \epsilon_T(i).
\]
The per-image query probability when using threshold $T$ is 
\[
f(T)=\mathbb{P}(X>T)=1-\Phi\!\Bigl(\frac{T-\mu}{\sigma}\Bigr),
\]
where $\Phi$ is the standard normal cumulative distribution function. By the mean value theorem, there exists some constant $c$ between $T_i$ and $T_i^*$ such that
\[
\bigl|f(T_i)-f(T_i^*)\bigr|\le \left|\frac{1}{\sigma}\phi\!\Bigl(\frac{c-\mu}{\sigma}\Bigr)\right|\cdot|T_i-T_i^*|\le \frac{1}{\sigma\sqrt{2\pi}}\epsilon_T(i).
\]
The last inequality is due to the function $\phi$, which is the standard normal pdf having the maximum value of $\frac{1}{\sqrt{2\pi}}$.
Now we define
\begin{align}\label{epsilon3}
\epsilon_3(i) = \frac{1}{\sigma\sqrt{2\pi}}\,\epsilon_T(i),
\end{align}
and we claim that if we denote
\[
p_i = f(T_i) = \mathbb{P}(X>T_i) \quad \text{and} \quad p^* = f(T_i^*) = \mathbb{P}(X>T_i^*),
\]
then it follows that
\begin{align}\label{p-pstar}
|p_i - p^*| \le \epsilon_3(i).
\end{align}
Next, we define the running query ratio i.e., the fraction of the samples that have been queried till that moment as
\[
R_N = \frac{1}{N}\sum_{i=1}^N I_i.
\]
Because the thresholds $T_i$ are computed from past data, the random variables $I_i$ are not independent, where we define $ I_i = \mathbf{1}\{X_i > T_i\}$. We let $\mathcal{F}_{i-1}$ denote the $\sigma$-algebra generated by $\{X_1,T_1,\dots,X_{i-1},T_{i-1}\}$. We now claim that the conditional expectation $\mathbb{E}[I_i\mid\mathcal{F}_{i-1}]$ is the predicted probability of querying the $i^\text{th}$ image, and by our earlier derivation \ref{p-pstar}, we have
\[
\Bigl|\mathbb{E}[I_i\mid\mathcal{F}_{i-1}] - p^*\Bigr| \le \epsilon_3(i).
\]
We now define the martingale differences
\[
D_i = I_i - \mathbb{E}[I_i\mid\mathcal{F}_{i-1}],
\]
so that $\mathbb{E}[D_i\mid\mathcal{F}_{i-1}] = 0$ and $|D_i| \le 1$. Setting
\[
M_N = \sum_{i=1}^N D_i,
\]
we obtain the decomposition
\[
R_N = \frac{1}{N}\sum_{i=1}^N \mathbb{E}[I_i\mid\mathcal{F}_{i-1}] + \frac{M_N}{N}.
\]
We now claim by the Azuma--Hoeffding inequality that for any $\epsilon>0$,
\begin{align}\label{azuma}
\mathbb{P}\Biggl(\Bigl|\frac{M_N}{N}\Bigr| \ge \epsilon\Biggr) \le 2\exp\Bigl(-\frac{N\epsilon^2}{2}\Bigr).
\end{align}
Moreover, since
\[
\left|\frac{1}{N}\sum_{i=1}^N \mathbb{E}[I_i\mid\mathcal{F}_{i-1}] - \overline{p^*}\right| \le \max_{1\le i\le N}\epsilon_3(i),
\]
following \ref{both_epsilon_together}, and \ref{azuma} we have 
\begin{align}\label{intermediate}
\mathbb{P}&\left(\left| R_N - \overline{p^*} \right| \le
\max_{1\le i\le N} \epsilon_3(i) + \epsilon\right) \ge\\ \nonumber
&1 - \delta_1 - 2\exp\left(-\frac{N\epsilon^2}{2}\right)
\end{align}
Since the ideal probability $p^*$ may vary with $i$, we define its average over $N$ images as $\overline{p^*}$,and we expect $R_N$ to concentrate around $\overline{p^*}$.
Our switching rule is as follows:
\begin{itemize}
    \item If $R_{i-1} < 0.075$, then we choose $z_i = z_{0.05}$, so that the ideal per-image query probability is $p^* = 0.05$.
    \item If $R_{i-1} \ge 0.075$, then we choose $z_i = z_{0.025}$, so that the ideal per-image query probability is $p^* = 0.025$.
\end{itemize}
We now define the sets
\[
A = \{ i \mid R_{i-1} < 0.075 \} \quad \text{and} \quad B = \{ i \mid R_{i-1} \ge 0.075 \},
\]
so that the system is in the ``$5\%$ regime'' when $i\in A$ and in the stricter ``$2.5\%$ regime'' when $i\in B$. The overall ideal average query probability is given by
\[
\overline{p^*} = \frac{1}{N}\sum_{i=1}^N p_i^* = \frac{|A|}{N}\times 0.05 + \frac{|B|}{N}\times 0.025.
\]
Now we define
\[
\alpha_N = \frac{|B|}{N}.
\]
Now we can write it in the form below since $|A|+|B|=N$
\[
\overline{p^*} = 0.05(1-\alpha_N) + 0.025\,\alpha_N = 0.05 - 0.025\,\alpha_N.
\]
Now our claim is that the stricter regime (where $p^* = 0.025$) cannot persist asymtotically for a nonzero fraction of time. For the sake of contradiction, that there exists a constant $\gamma > 0$ and infinitely many $N$ such that
\[
\alpha_N = \frac{|B|}{N} \ge \gamma.
\]
Then the overall ideal probability satisfies
\[
\overline{p^*} \le 0.05 - 0.025\,\gamma.
\]
By the concentration bound established above, for sufficiently large $N$, following \ref{intermediate} we have
\[
R_N \approx \overline{p^*} \le 0.05 - 0.025\,\gamma + o(1) < 0.075.
\]
Since for sufficiently large N, we assume $(\max_{1\le i\le N}\epsilon_3(i) + \epsilon )= o(1).$
But if $R_N < 0.075$, then by the switching rule the system will revert to the $5\%$ regime in subsequent steps. This contradiction shows that it is unsustainable for $\alpha_N$ to be bounded away from zero; hence, we have
\begin{align}\label{zero}
\limsup_{N\to\infty} \frac{|B|}{N} = 0.
\end{align}
Now we define
\[
F_N = \Bigl|\frac{1}{N}\sum_{i=1}^N p_i^* - 0.05\Bigr| = 0.025\,\frac{|B|}{N},
\]
and we claim that $F_N \to 0$ as $N\to\infty$. Thus, incorporating the switching rule, we obtain the overall deviation bound
\[
|R_N - 0.05| \le F_N + \max_{1\le i\le N}\epsilon_3(i) + \epsilon,
\]
which holds with high probability. Since both $\max_{1\le i\le N}\epsilon_3(i)$ and $F_N$ vanish as $N\to\infty$, it follows that
\[
\lim_{N\to\infty} \mathbb{P}\Bigl(|R_N - 0.05| \ge \epsilon\Bigr) = 0 \quad \text{for all } \epsilon > 0.
\]